\documentclass[conference]{lib/IEEEtran}
\pdfoutput=1

\usepackage{amsmath}
\usepackage{amssymb}
\usepackage{booktabs,siunitx}
\usepackage{arydshln}
\usepackage{times}
\usepackage{etoolbox}
\usepackage[export]{adjustbox}
\usepackage{framed}
\usepackage{color, colortbl}
\usepackage{setspace}
\usepackage{multirow}
\usepackage{tabularx}
\usepackage{tikz}
\usepackage{placeins}
\usepackage{cuted} 
\usepackage{caption} 
\usepackage{pifont}

\usepackage{enumitem}
\usepackage{bbm}
\usepackage{dsfont}
\usepackage{algorithm}
\usepackage{algpseudocode}
\usepackage{graphicx}

\usepackage[colorlinks,allcolors=black,pdftex]{hyperref}

\usepackage[capitalize]{cleveref}
\crefname{section}{Sec.}{Secs.}
\Crefname{section}{Section}{Sections}
\Crefname{table}{Table}{Tables}
\crefname{table}{Tab.}{Tabs.}

\definecolor{rowzebra}{HTML}{f5f7fa}
\begin{document}
    \title{Disentangling Racial Phenotypes:\\Fine-Grained Control of Race-related Facial Phenotype Characteristics}
\author{
\begin{tabular}{@{}c@{}}
    Seyma Yucer$^1$, 
    Amir Atapour Abarghouei$^1$,
    Noura Al Moubayed$^1$,
    Toby P. Breckon$^{1,2}$
\end{tabular}
\\Department of \{$^1$Computer Science $\mid$ $^2$Engineering\}, Durham University,
Durham, UK\\
}

\maketitle

\begin{strip}
\vspace{-1.3cm}
\centering
\includegraphics[width=\textwidth]{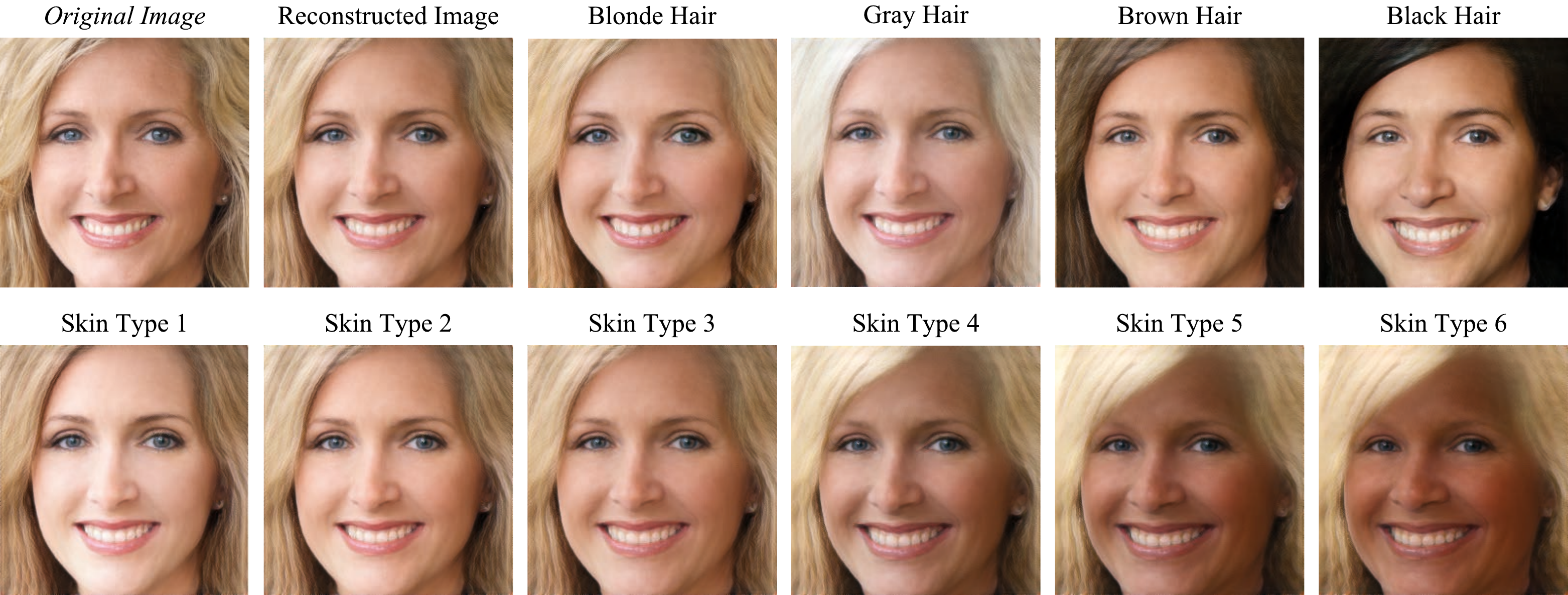}
\captionof{figure}{Generated images with controlled race-related phenotypes by our proposed framework.}
\label{fig:front}
\end{strip}
\begin{abstract}


Achieving an effective fine-grained appearance variation over 2D facial images, whilst preserving facial identity, is a challenging task due to the high complexity and entanglement of common 2D facial feature encoding spaces. Despite these challenges, such fine-grained control, by way of disentanglement is a crucial enabler for data-driven racial bias mitigation strategies across multiple automated facial analysis tasks, as it allows to analyse, characterise and synthesise human facial diversity. In this paper, we propose a novel GAN framework to enable fine-grained control over individual race-related phenotype attributes of the facial images. Our framework factors the latent (feature) space into elements that correspond to race-related facial phenotype representations, thereby separating phenotype aspects (e.g. skin, hair colour, nose, eye, mouth shapes), which are notoriously difficult to annotate robustly in real-world facial data. Concurrently, we also introduce a high quality augmented, diverse 2D face image dataset drawn from CelebA-HQ for GAN training. Unlike prior work, our framework only relies upon 2D imagery and related parameters to achieve state-of-the-art individual control over race-related phenotype attributes with improved photo-realistic output. 


\end{abstract}

    \section{Introduction}
\noindent

Analysing and characterising human facial diversity is crucial for automated facial analysis tasks, especially as increasing research reveals the presence of racial bias causing disparate performances for racial groups \cite{garcia2019harms,yucer2020exploring,georgopoulos2021mitigating}. 
Moreover, recent studies \cite{yucer2022measuring,yucer2023racial,merler2019diversity} highlight the advantage of race-related facial attribute level analysis of racial bias to avoid using ill-defined racial categories and further specify the race-related facial phenotype attribute categories for racial bias evaluation.

On the other hand, disentanglement learning, with its primary objective being to capture independent data variation factors, shows promise for achieving group fairness/demographic parity \cite{locatello2019fairness} for classification tasks and can be particularly relevant in mitigating racial bias. Earlier studies \cite{creager2019flexibly, locatello2019fairness} discuss how disentangled representation learning can enhance group fairness by isolating variations into independent components, thereby improving interpretability, and simplifying downstream prediction tasks. In contrast to conventional image-to-image transition methods \cite{choi2020stargan}, where the aim is learning a mapping between different visual domains, disentanglement learning aims to isolate such independent components of data to enable explicit control on the generated images.


Consequently, in this study, we aim to explicitly control race-related facial phenotype attributes, setting the foundation for creating controlled face image variations for future potential solutions to mitigate racial bias within automated facial analysis tasks. Most pertinent to our research, ConfigNet \cite{kowalski2020config} provides a framework using HoloGAN \cite{nguyen2019hologan} for parametric rendering over 2D facial images by incorporating 3D parameters from synthetic data. The objective of ConfigNet \cite{kowalski2020config} is to generate realistic and controllable face images via modelling and generating of intricate attribute parameters (not present in  the 2D dataset) within a 3D synthetic image dataset, bridging the gap between neural rendering and traditional rendering pipeline. Our aim of is specifically related with its ability to render both complex, multiple identity-relevant and -irrelevant factors into the latent space. Yet, instead of utilising 3D synthetic data, we derive the parameters in a 2D image space, which is significantly more challenging but yet has greater real-world applicability. We aim to have realistic image generation with controllable identity-relevant attributes in a factorised latent space.

To this end, we develop an enhanced framework, solely grounded on 2D imagery and its metric-based parameters, for controlling specific race-related facial phenotypes such as skin and hair colour, and shapes of nose, eyes, and mouth. Our approach emphasises explicit control over these facial parameters, which are delineated and quantified using 2D image evaluations. Initially, we define these race-related phenotype parameters through 2D metric-based evaluations, subsequently factorised them into the latent space. We then improve the ConfigNet framework by adopting the generator-discriminator architecture of StyleGAN2, replace the synthetic data and its 3D parameters in favour of 2D high-resolution training data for which we curate an augmented, diverse dataset derived from CelebHQ. In this paper, our key contributions are as follows:

\begin{figure*}[ht]
\begin{center}
\includegraphics[width=\textwidth]{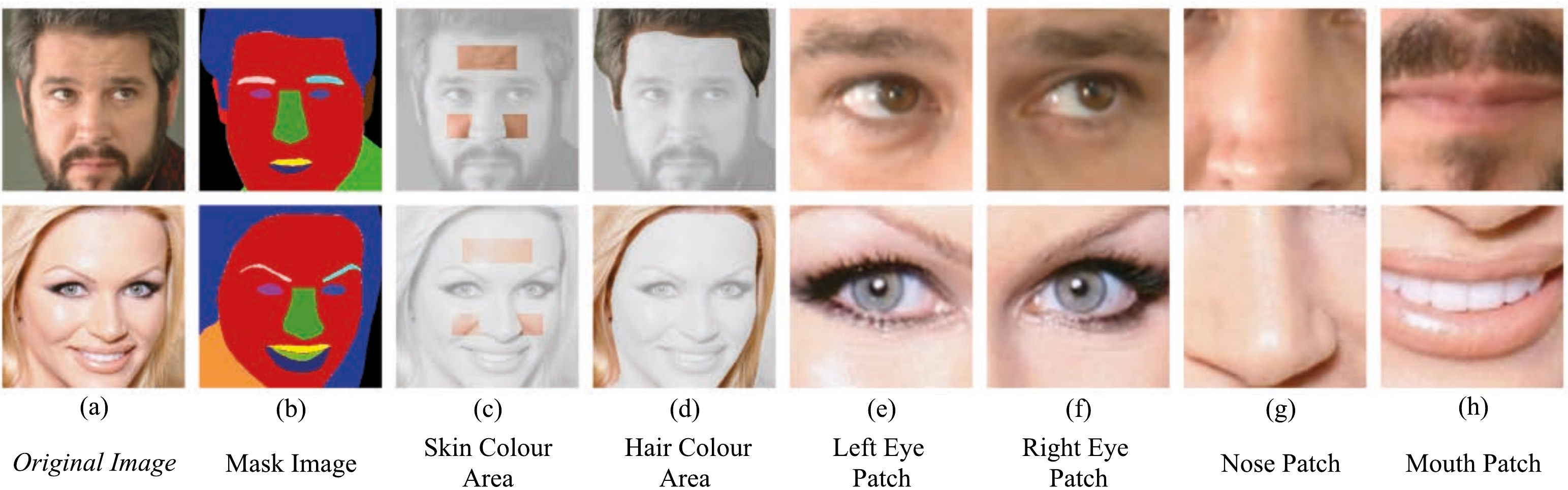}
\end{center}
\caption{Metric-based parameters for race-related facial phenotypes: (a) Top column images are sourced from CelebA-HQ \cite{karras2017progressive}, (b) Mask images provided by MaskGAN \cite{CelebAMask-HQ}. (c) The facial skin area used for skin colour and (d) the hair area used for hair colour. (e-h) The specific face patch inputs applied for feature extraction. } 
\label{fig:paremeters}
\end{figure*}

\noindent
\begin{itemize}[leftmargin=*]

\item We propose a framework that achieves explicit control over identity-relevant race-related facial phenotypes via a single factorised and disentangled latent space.

\item Our framework relies on simple hand-crafted 2D metrics parameters obtained by public face dataset, eliminating the need for 3D render data or manual auditing.

\item We introduce the CelebA-HQ-Augmented-Cleaned dataset, which is the first semi-synthesised, manually-cleaned, high-quality dataset encompassing over 26,500 images with a diverse distribution.

\item We demonstrate that our proposed framework achieves both higher image quality and controllability on race-related facial phenotype attributes in comparison to \cite{kowalski2020config}.
\end{itemize}

\section{Related Work}
\noindent
\textbf{Controllable GANs:} The latest advancements in Generative Adversarial Networks (GAN) \cite{karras2020analyzing,zhang2022styleswin} not only enable high-quality face image generation but also provide control and editing capabilities within the image generation process \cite{kowalski2020config,deng2020disentangled}. Whilst, many common controllable image-to-image-based  \cite{kim2022style,lee2020byeglassesgan,choi2020stargan} and latent space interpolation-based methods \cite{he2018facial,abdal2020image2stylegan} offer ways to control facial attributes, 
they do not inherently offer a factorised latent space for explicit control over image attributes.

Existing literature on controllable GAN is separated into two categories following \cite{shoshan2021gan}: relative control \cite{
shen2018faceid,harkonen2020ganspace,shen2020interpreting,balakrishnan2021towards} and explicit control \cite{kowalski2020config,deng2020disentangled,tewari2020stylerig,shoshan2021gan}. Relative control provides basic manipulations like changing illumination or facial rotation, whilst explicit control enable precise manipulations, such as setting the illumination to a lighter shade or rotating the face by exact angles (e.g. $30^{\circ}$ to the left).

A widely adopted approach for both relative and explicit control of images within generative process is based on identifying disentanglement properties in the latent space corresponding to image attributes \cite{kowalski2020config,deng2020disentangled,tewari2020stylerig,shoshan2021gan}. Numerous studies \cite{nickabadi2022comprehensive,xu2021faceshapegene,lee2020byeglassesgan} have identified such facial attribute properties, such as head pose, lighting, facial expressions, facial accessories, gender, and age, aiming to effectively disentangle such attributes from the facial identity. Such facial attributes can be categorised as either identity-relevant or identity-irrelevant \cite{nickabadi2022comprehensive}. 
Identity-relevant attributes, such as racial features such as nose and eye shapes, define distinctive facial characteristics that remain same under different expressions and poses. Conversely, identity-irrelevant attributes such as smiling or head pose are non-distinctive, as any alterations to them do not impact the overall identity. Accordingly, we aim to control explicitly identity-relevant race-related facial phenotypes attributes such as skin and hair colour, and shapes of nose, eyes, and mouth proposed by \cite{yucer2022measuring}.

\textbf{Disentanglement via GANs:} Moreover, disentangling identity-relevant attributes is more complex task than controlling image due to their higher mutual information with facial identity, compared to identity-irrelevant attributes. Yet, much of the existing disentanglement literature primarily addresses identity-irrelevant attributes including head pose, expressions, mouth openness, smiling, and makeup \cite{lee2020byeglassesgan,gu2019ladn,wang2019dft}. 
For example, StyleRig \cite{tewari2020stylerig} provides fine-grained control over facial images generated by StyleGAN, integrating an additional layer that captures 3D pose and expression variations. More recently, \cite{pang2023dpe} proposes a novel self-supervised disentanglement framework to decouple pose and expression without using 3DMMs and paired data. However, despite this progress in GAN, achieving explicit control on identity-relevant facial attributes over the generative process remains a challenge. Such explicit control requires not only keeping photo-realism and facial identity but also changing the single individual attribute in a desired way. Consequently, 3D face representations in generative models, such as 3DMM or equivalent 3D meshes, provide a deeper level of control in the latent space \cite{nguyen2019hologan,chan2021pi,xia2022survey}. While it can facilitate disentanglement by leveraging depth and shape information, obtaining an accurate and detailed 3D imagery and supervision (attribute labels and representations) is challenging and furthermore such high-fidelity 3D imagery makes GAN training even more complex and computationally intensive \cite{xia2022survey}.

Consequently, in this study, we achieve explicit control over race-related facial phenotype parameters solely through the use of 2D metric-based evaluations. Furthermore, we introduce the CelebA-HQ-Augmented-Cleaned dataset contains semi-synthesised, diverse, manually-cleaned high-quality images. Additionally, we propose an enhanced version of the Confignet \cite{kowalski2020config} framework that integrates StyleGAN \cite{karras2019style} and eliminates the requirement for 3D rendering parameters.

    \section{Methodology}
\noindent
Our method employs two 2D face image datasets: a supervised set \(I_{C}\) sampled from CelebA-HQ \cite{karras2017progressive} and an unsupervised set \(I_F\) from FFHQ \cite{karras2019style}. The primary distinction between \(I_{C}\) and \(I_F\) is their intended use. \(I_{C}\) introduces race-related facial phenotype attributes into the factorised latent space, while \(I_F\) is used without any paired supervision  (facial phenotype attribute). Our framework does not require any supervision during the test phase. We detail the process of acquiring race-related facial phenotype attributes of \(I_{C}\) to factorise in latent space in Section \ref{sec:2:1} and further explain our framework in Section \ref{sec:2:2}.

\begin{table*}[ht]
\centering
\caption{Dimensions and descriptions of race-related facial phenotype attributes in factorised latent space.}
\begin{tabular}{@{}llll@{}}
\toprule
\textbf{Phenotype} &
  \textbf{Representation} &
  \textbf{Description} &
  $Input \rightarrow Output$ \\ \midrule
Skin Colour & \( \theta_{skin} =\{ V_{mean},  S_{mean},  Cr_{mean} \}\)  & Melanin, Greyness,  Redness & $\mathbb{R}^{3} \rightarrow \mathbb{R}^{3}$\\
Hair Colour & 
\( \theta_{hair} =\{ V_{mean}, S_{mean},  Cr_{mean} \}\) 
& Melanin, Greyness,  Redness & $\mathbb{R}^{3} \rightarrow \mathbb{R}^{3}$  \\
Left Eye & \(\theta_{lefteye}=\{q_1,q_2...,q_{125}\}\) & Left eye  feature vector  &$\mathbb{R}^{125} \rightarrow \mathbb{R}^{125}$\\
Right Eye &\(\theta_{righteye}=\{q_1,q_2...,q_{125}\}\)&Right eye  feature vector  & $\mathbb{R}^{125} \rightarrow \mathbb{R}^{125}$ \\
Nose &\(\theta_{nose}=\{q_1,q_2...,q_{128}\}\) &Nose feature vector &  $\mathbb{R}^{128} \rightarrow \mathbb{R}^{125}$\\
Mouth (Lips) & \(\theta_{mouth}=\{q_1,q_2...,q_{128}\}\)& Mouth feature vector & $\mathbb{R}^{128} \rightarrow \mathbb{R}^{125}$\\ 
\bottomrule
\end{tabular}

\label{tab:phenotypes}
\end{table*}

\subsection{Race-related Facial Phenotypes in Factorised Latent Space} \label{sec:2:1}
\noindent
Prior work \cite{yucer2022measuring} identifies a set of observable race-related facial phenotype characteristics that are specific to face and correlated to the racial profile of the subject. These representative race-related facial attributes encompass skin, hair colour, eye, nose, and lip shape. We use the same attribute categories within the factorised latent space and denote each of them with \(\theta\) corresponding naming as in Table \ref{tab:phenotypes}. As a result, each facial image within the supervised dataset contains various predetermined facial phenotype: skin colour \(\theta_{skin}\), hair colour \(\theta_{hair}\), nose \(\theta_{nose}\), eye \(\theta_{left\_eye}\) and \(\theta_{right\_eye}\), and mouth \(\theta_{mouth}\) features. We derive hand-crafted metric-driven representations for these specific phenotype attributes, avoiding subjective annotations. Following this, akin to the methodology in ConfigNet \cite{kowalski2020config}, each phenotype attribute is factorised into $k$ components $\theta_1$ to $\theta_k$, as follows:

\begin{equation}
 \theta \in \mathbb{R}^m =  \mathbb{R}^{m_1} \times \mathbb{R}^{m_2} \times \dots \times \mathbb{R}^{m_k}
\end{equation}
\\
\noindent
Each $\theta_i$ corresponds to a semantically meaningful facial phenotype attribute to generate $I_{C}$. The supervised data encoder $E_C$ maps each $\theta_i$ to $z_i$, a part of $z$, which thus factorises $z$ into $k$ parts. The factorised latent space enables manipulation of pre-defined attributes in generated images by swapping specific attributes such as skin colour of the part represented by $z_i = E_{C_i}(\theta_i)$. We also present such attributes and descriptions in Table \ref{tab:phenotypes}.

\noindent
\textbf{Skin and Hair Colour:} We utilise skin and hair segmentation masks on face images in order to quantify skin and hair colour. MaskGAN \cite{CelebAMask-HQ} provides hand-annotated mask images (as shown in the second column (b) of Figure \ref{fig:paremeters}.) for CelebA-HQ \cite{karras2017progressive} dataset with 19 classes including all facial components and accessories. We restrict the skin region on the skin segments via facial landmark points, considering the potential overlap of beard and eyeglasses on the face. Subsequently, we measure the melanin, greyness, and redness values within the selected skin region and the hair region (column (c) for skin and (d) for hair in Figure \ref{fig:paremeters}). As a baseline for our work, ConfigNet \cite{kowalski2020config} employs these values for hair colour analysis using a 3D image rendering software. Instead, we estimate the 2D colour spaces of the skin and hair regions to capture the \textit{melanin, greyness, and redness} values within these regions. Specifically, for the \textit{melanin} representation, we convert the skin and hair pixels (separately) from the RGB colour space to the HSV colour space and measure the mean value of the $(V)$ channel describing the intensity of the colour. Increased $(V)$ corresponds to a lighter skin tone due to decreased melanin levels, with reverse correlation providing skin colour representation. Similarly, to assess the \textit{greyness} representation, we estimate the mean saturation value $(S)$ from the HSV space, which represent the degree of greyness. Lastly, we convert the RGB colour space to the YCrCb colour space and extract the $(Cr)$ channel mean value within the selected skin and hair regions to capture the redness component. 
\\
\noindent
\textbf{Nose, Lip, Eye Shape Feature} Furthermore, to extract representations of the eyes, nose, and mouth from images, we produce 64$\times$64 pixel patch images, as shown in Figure columns (e-g) \ref{fig:paremeters} using facial landmarks. For each facial region (left eye, right eye, lips, and mouth), we train individual MobilenetV2 networks \cite{sandler2018mobilenetv2} using the original CelebA dataset and its facial attribute categories excluding CelebA-HQ \cite{karras2017progressive} samples to be later utilised as $I_C$. Features are then extracted from the final layer of corresponding model. As prior work \cite{yucer2022measuring} also categorises the eyes, nose, and mouth into two groups, we utilise the ground truth labels from CelebA attributes: ``Big Nose" for the nose patch, ``Big Lips" for the mouth patch, and ``Narrow Eyes" for both left and right eye patch images.

\begin{figure*}[ht]
\begin{center}
\includegraphics[width=\textwidth]{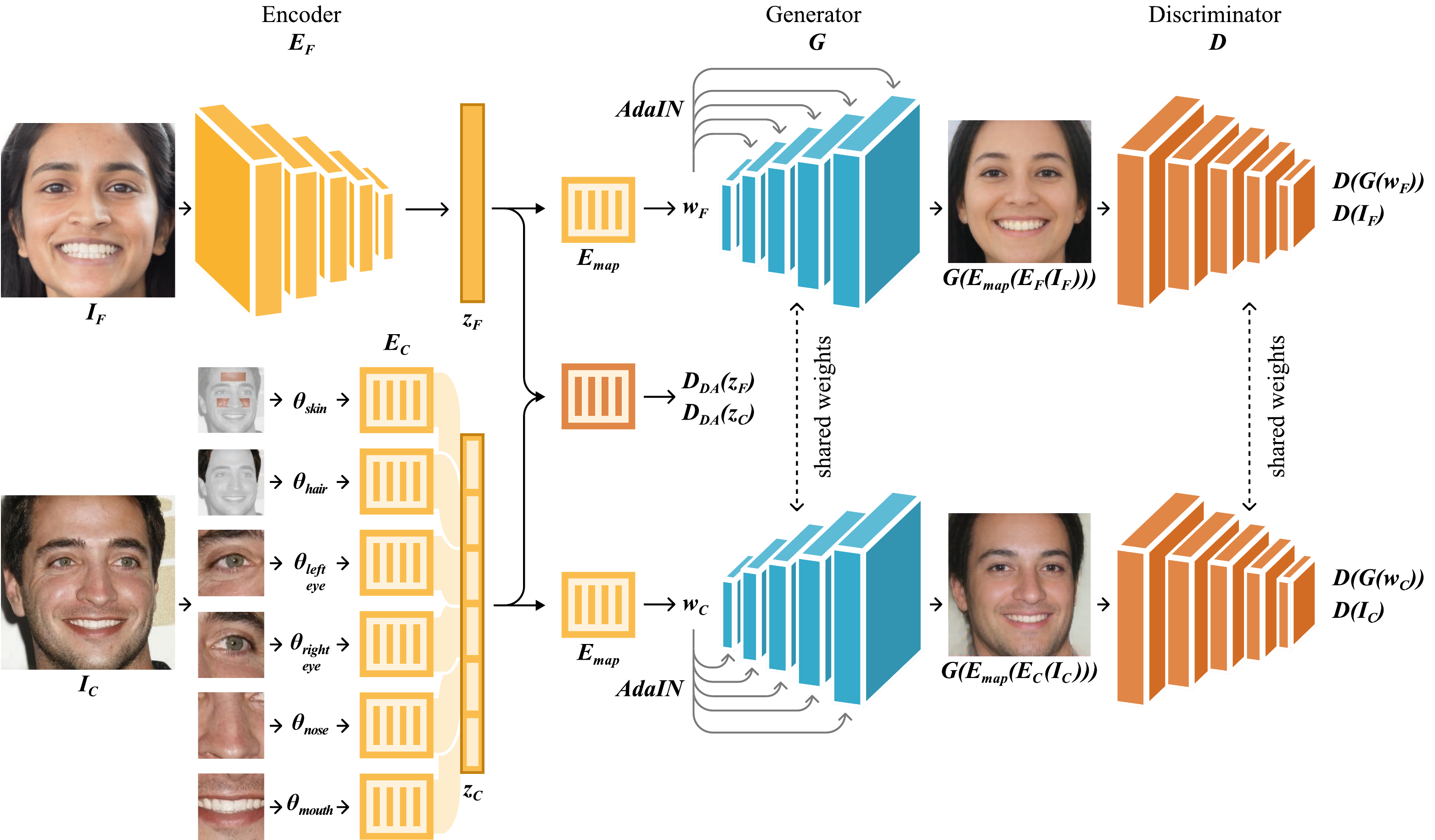}
\end{center}
\caption{The proposed framework employs two encoders $E_F$ and $E_C$ that encode face images $I_F$ and $I_{C}$ in latent space vectors $z_F$ and $z_C$, respectively. These vectors are further mapped into $w_F$ and $w_C$ using $E_{map}$, which are then fed into the shared decoder $G$ for image generation. A domain discriminator \(D_{DA}\) ensures the similarity of latent distributions generated by $E_F$ and $E_C$.} 
\label{fig:overview}
\end{figure*}

\begin{figure}[b]
 \centering
 \includegraphics[width=\columnwidth]{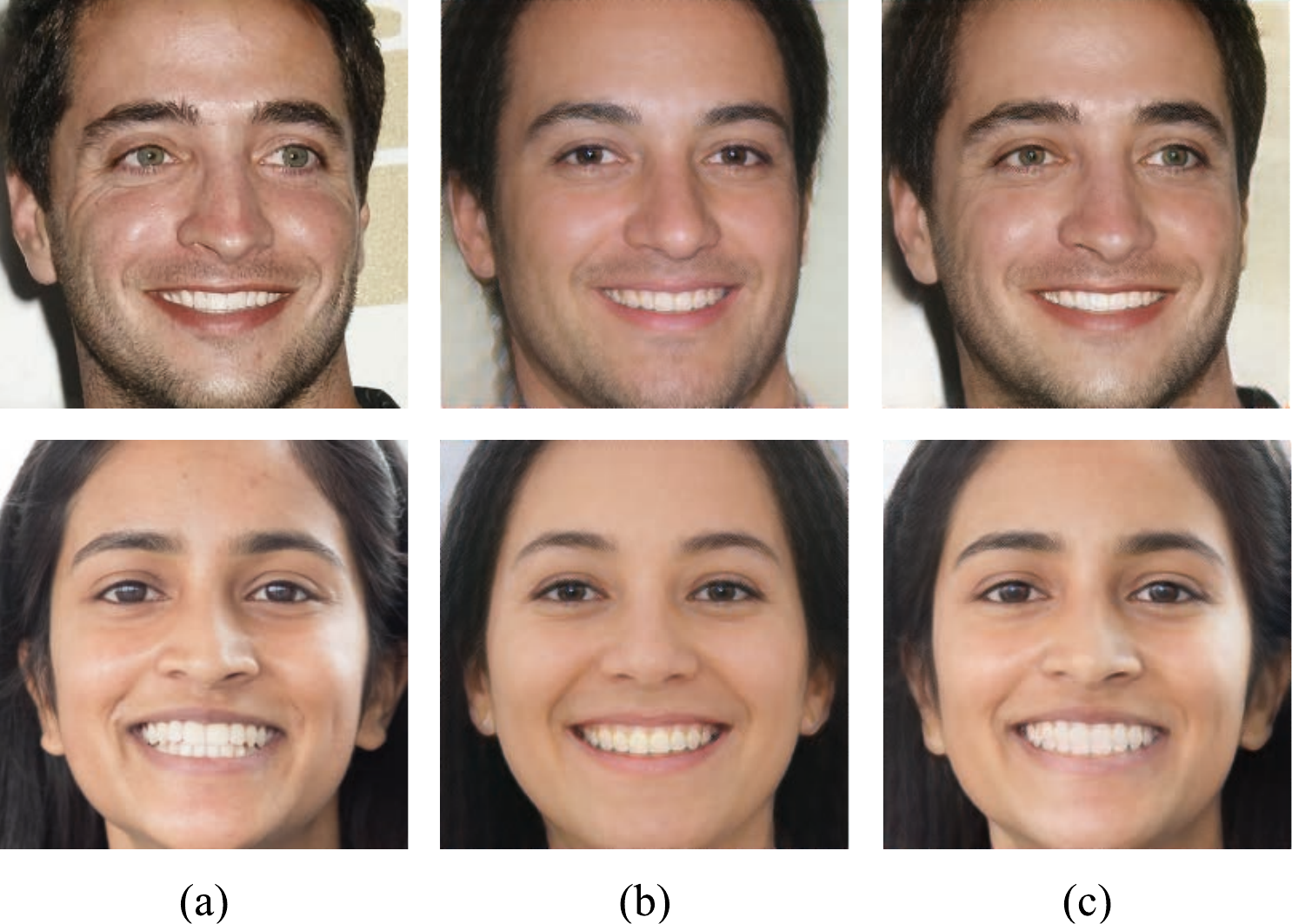}
 \caption{The impact of one-shot learning through fine-tuning. (a) Original image. (b) Reconstructed image after second-stage training. (c) Reconstructed image after fine-tuning}
 \label{fig:finetune}

\end{figure}

\subsection{Proposed Framework} \label{sec:2:2}
\noindent
Building on the structure of the baseline \cite{kowalski2020config}, our method incorporates a decoder \(G\) and two encoders, \(E_F\) and \(E_C\) and a discriminator \(D\) as can be seen in Figure \ref{fig:overview}. \(E_F\) is a ResNet-50 backbone architecture \cite{he2016deep} pre-trained on ImageNet \cite{russakovsky2015imagenet}. \(E_C\) is a set of separate multi-layer perceptrons (MLPs) \(E_{C_i}\) for each of the corresponding \(\theta{i}\) in Table \ref{tab:phenotypes}. These encoders \(E_C\) and \(E_F\) embed both \(I_F\) and \(I_C\) into a unified factorised latent space \(z_F\) and \(z_C\) respectively.Unsupervised set \(I_F\) is provided to its encoder as images from the set \(I_F\), whereas supervised data is represented as vectors \(\theta \in R^m\), which thoroughly delineate the content of the associated image in \(I_{C}\) (as explained in Section \ref{sec:2:1}). Subsequently, both \(z_F\) and \(z_C\) are transformed into \(w_F\) and \(w_C\) using the StyleGAN2 mapping network \(E_{\text{map}}\), which comprises eight fully-connected layers. The vector size of \(z_F\), \(z_C\) and \(w\) are all 512.

Whilst the baseline work \cite{kowalski2020config} employs separate discriminator networks, \(D_F\) and \(D_C\), for both real and synthetic data to enhance image realism, we implement a shared discriminator \(D\) in the second stage, given our sole dependence on 2D image sets, negating the need to close the realism gap caused by the use of synthetic data in \cite{kowalski2020config}. Similar to \cite{karras2019style}, we apply a two-stage training strategy. 
\\
\noindent
In the first stage, we train a shared StyleGan2  generator \(G\) with its mapping encoder \(E_{map}\) \cite{karras2019style}, and separate discriminators \(D_F\) and \(D_C\) and encoder \(E_C\). $z_F$ is sampled from the normal distribution and encoder \(E_F\) is not included in this stage. With the combined StyleGAN2 architecture \cite{karras2019style}, the first stage loss is:

\begin{equation}
\begin{split}
L_1  &= L_{GAN_G}(D_F,G(w_F)) + L_{GAN_G}(D_{DA},z_C) \\ & + L_{GAN_G} (D_C,G(w_C))+ \lambda_{perc} L_{perc}(G(w_C),I_{C})
\end{split}
\end{equation}
where $
L_{GAN_G}(D, x) = -\log(D(x)).
$ As StyleGAN maps the input latent vector $z$ to an intermediate latent space $w$, we first map factorised latent space $z_C$ to $w_C$ and then control the generator through adaptive instance normalisation (AdaIN) at each convolution layer of $G$. 
We remove eye loss and identity loss as we do not observe any improvement after adopting StyleGAN2. Following \cite{kowalski2020config}, we set the same loss weights as follows: domain adversarial loss weight $\lambda_{DA} = 5$, gradient penalty loss weight $\lambda_{R1}=10$, perceptual loss weight in the first stage ${\lambda}_{perc}=0.00005$. The adversarial losses on the images including the style generator and discriminator losses are equally weighted.
\\
\noindent
In the second stage, we introduce $E_F$ and a single shared discriminator $D$, where the pre-trained weights of $D_F$ are utilised for training $D$.
The second stage loss is:

\noindent
\begin{equation}
\begin{split}
L_2  &= L_1+ \lambda_{perc} L_{perc}(G(w_F),I_F) + \log{(1-D_{DA}(z_F)) }
\end{split}
\end{equation}
where the aim of \( \log(1 - D_{DA}(z_F)) \) is to align the output distribution of $E_F$
with that of $E_C$. We set perceptual loss weight ${\lambda}_{perc}=10$ in this stage. In our experiments, the two-stage training enhanced both controllability and image quality, while attempts to single-stage training process (training all encoders, the generator, discriminator collectively in one iteration) result in unsatisfactory image generation.
\\
\noindent
\textbf{One-shot learning by fine-tuning:} 
Following the approach in \cite{kowalski2020config}, we employ a one-shot learning procedure to reduce the identity gap by fine-tuning the generator using individual images. This identity gap between the original and reconstructed images as well as improved reconstruction achieved in this stage are presented in Figure \ref{fig:finetune}. In a similar vein, we fine-tune our generator on \(I_F\) by minimising the subsequent loss:

\begin{figure}
\centering
\includegraphics[width=1.1\linewidth,page=1]{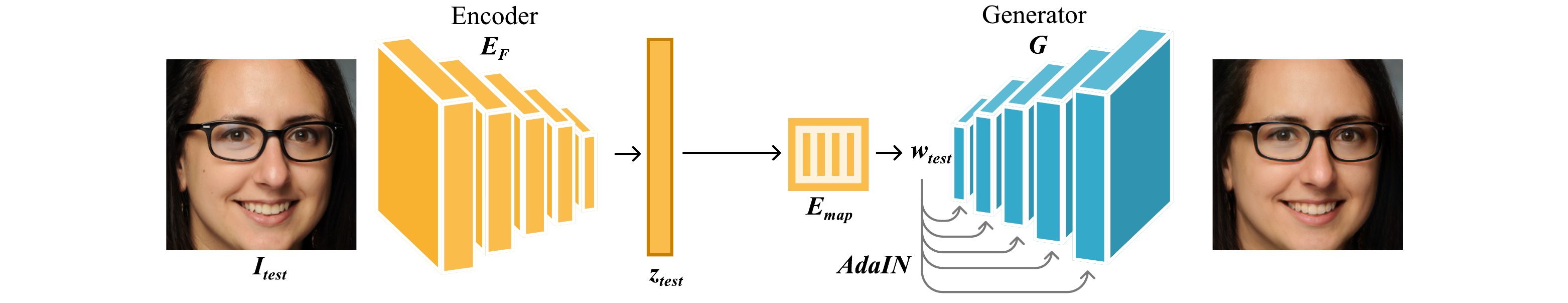}
\caption{Inference of our proposed framework.}
    \label{fig:inference}
\end{figure}

\begin{figure*}[ht]
\begin{center}
\includegraphics[width=\textwidth]{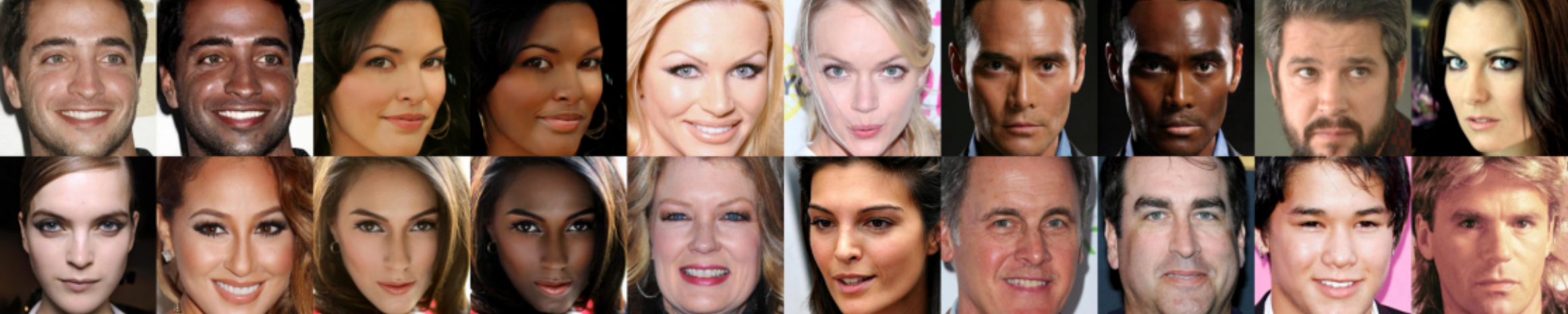}
\end{center}
\caption{A selection of images from CelebA-HQ-Clean-Augmented. While some images are augmented using the method proposed by \cite{yucer2020exploring}, others, both original and augmented, are removed due to low imaging conditions and pose discrepancies.}
\label{fig:celebahq_samples}
\end{figure*}

\begin{equation}
\begin{split}
L_{ft}  &= L_{GAN_G}(D,G(\hat{w_F})) 
+ \log{(1-D_{DA}(\hat{z_F})) }
 \\ & +  L_{perc}(G(\hat{w_F}),I_F) + L_{face}(G(\hat{w_F}),I_F)],
\end{split}
\end{equation} where \(L_{face}\) is a perceptual loss with VGGFace \cite{parkhi2015deep} as the pre-trained network.
We optimise over \(G\) as well as \(z_F\) which is initialised with \(E_F(I_F)\).
The addition of a \(L_{face}\) improves the perceptual quality of the generated face images, whilst it is not noticeable during the main training phase, since fine-tuning lacks the regularisation achieved through  training on a large number of images.
\\
\noindent 
\textbf{Fine-grained Phenotype Control}
To have fine-grained control over the latent space generated by \(E_F\), we adopt the gradient descent-based minimisation algorithm presented by \cite{kowalski2020config}. This enables targeted modifications, such as adjusting skin colour or hair colour darkness level, while ensuring the rest of the facial attributes remain the same (for a detailed description, see \cite{kowalski2020config}).

\noindent
\textbf{Inference} We present the inference pipeline of our framework in Figure \ref{fig:inference}. Importantly, our approach achieves disentanglement of race-related facial phenotypes without requiring additional attribute labels or representations. This is achieved through the training of $E_F$, which encodes these phenotypes within a factorised latent vector space utilised by the Generator $G$. For any given 2D image $I_{\text{test}}$, it is encoded by $E_F$ and $E_{\text{map}}$ in sequence, and then reconstructed by $G$. Simultaneously, the control of the generated image is enabled by modifying specific components of $z_{\text{test}}$.

    \section{Experimental Results}
\noindent
In this section, we explain our training setup and experimental results to evaluate photorealism and controllibility.

\subsection{Datasets}
\noindent
We utilise the FFHQ \cite{karras2019style} and CelebA-HQ datasets for training of our framework. 
FFHQ dataset \cite{karras2019style} contains 60,000 high-resolution images of size 1024$\times$1024 pixels. We utilise 50,000 samples from FFHQ for our training set as our primary source of unsupervised images \(I_F\) and the same 10,000 samples for the validation set ($I_{test}$) for a consistent comparison of results with ConfigNet \cite{kowalski2020config}. CelebA-HQ, a subset of CelebA, offers 30,000 high-resolution images, each at a resolution of 1024$\times$1024 \cite{karras2017progressive} and is the source of CelebA-HQ-Clean-Augmented (supervised set, \(I_C\)).
\\
These datasets consist of an imbalanced racial distribution. For instance, \cite{maluleke2022studying} reveals that the FFHQ dataset consists of 69\% White, 4\% African, and 27\% individuals who are neither African nor white. Similarly, \cite{zhu2022celebv} indicates that CelebA-HQ contains over 70\% White individuals and fewer than 10\% of African. To address this, we introduce CelebA-HQ-Clean-Augmented which is a semi-augmented high-quality image set. We align all the face images from those datasets to a standard reference frame using landmarks from OpenFace \cite{baltrusaitis2018openface} and reduce the resolution to 256$\times$256 pixels.

\noindent
\textbf{CelebA-HQ-Clean-Augmented}: To address the lack of diversity within the GAN training dataset, we apply a prior adversarial data augmentation technique to facilitate the transfer of race-specific facial features \cite{yucer2020exploring}. From the original 30,000 CelebA-HQ images, we augmented another 30,000 images by transferring all the images from the Caucasian to the African domain. However, both the original and synthesised images exhibit poor imaging conditions and not all of the original images actually belong to Caucasian subjects, which may cause faulty or erroneous parameter estimation.
Moreover, as skin colour estimation relies on colour spaces, we prioritised images without prominent shading or lighting that may mislead the skin colour evaluation. Accordingly, we manually clean and select a refined dataset containing 26,513 images; 17,861 original and 8,652 augmented. Figure \ref{fig:celebahq_samples} shows exemplar images from the curated CelebA-HQ-Clean-Augmented dataset.

\subsection{Image Quality - Photorealism}
\begin{figure*}[ht]
\begin{center}
\includegraphics[width=\textwidth]{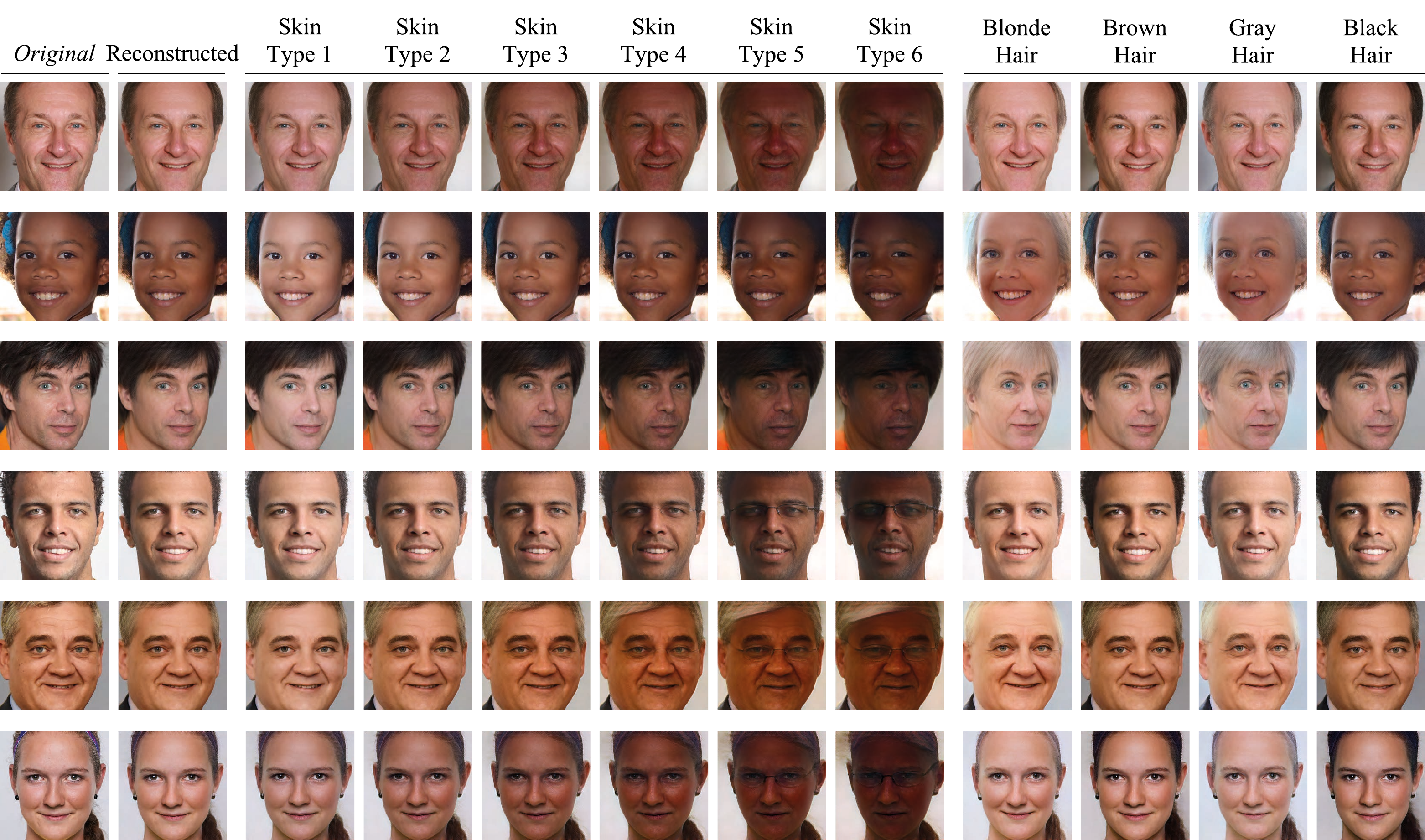}
\end{center}
\caption{Generated and controlled images from $G(E_{map}(E_{F}(I_{test}))$. From the top row to the following rows, the sequence respectively shows original and reconstructed images, followed by generated images with associated attribute changes. We modify the corresponding index of $z_{test}=(E_{F}(I_{test}))$ to synthesise attribute-modified images.
} 
\label{fig:generateds}
\end{figure*}

\noindent
In Table \ref{tab:fid}, we measure the photorealism of our generated images using the Frechet Inception Distance (FID) \cite{heusel2017gans} and compare our results with ConfigNet \cite{kowalski2020config}. First, we examine the FID score between the FFHQ and our CelebA-HQ-Clean-Augmented dataset. Since ConfigNet \cite{kowalski2020config} utilises raw synthetic images, the SynthFace dataset, there is a noticeable feature distance when compared to FFHQ. By replacing SynthFace dataset with CelebA-HQ-Clean-Augmented face dataset, we not only eliminate the need for synthetic data but also significantly improve the distribution difference of training sets by lowering FID score by 12 points (from 52,19 to 40,81 $\downarrow$). 
In the subsequent evaluation, we test the FID performance of the first stage by generating random images from the first-stage trained generator \(G\). Notably, our framework achieves a lower perceptual distance score, indicating higher image quality and more realistic image generation. Subsequently, we show our second-stage trained model reconstruction quality using \(E_F\), we re-generate FFHQ evaluation set, $I_{test}$, and calculate FID score between $G(E_{map}(E_F(I_{test}))$ and $I_{test}$. Our approach consistently produces more realistic images compared to \cite{kowalski2020config}. 
\\
\noindent
Additionally, we modify the relevant attribute index location of the latent space vector $z_F=E_F(I_{test})$ to control the skin and hair colour of the generated image while preserving the other features. As a result, we present qualitative results for our generated images, encompassing both reconstructed and manipulated images with focused attribute variations in Figure \ref{fig:generateds}.

\begin{table}[b]
\caption{FID score for FFHQ, CelebA-HQ-Clean-Augmented, and images obtained with our decoder $G$ and latent vectors $z_F$ from the real-image encoder $E_F$.}
\centering
\begin{tabular}{@{}lcc@{}}
\toprule
\textbf{Method} & \begin{tabular}[c]{@{}c@{}}\textbf{ConfigNet}\\ \cite{kowalski2020config}\end{tabular} & \textbf{Ours} \\ 
\midrule
$I_{C}$      &   52.19                   &         40.81\\ 
$G(z), z \approx N(0,(I))$ no 2nd stage      &        43.05         &  39.55   \\ 

$G(E_F(I_F))$           &   33.41                        &  28.64             \\
\end{tabular}

\label{tab:fid}
\end{table}

\subsection{Controllability}
\noindent
We adopt the ConfigNet \cite{kowalski2020config} controllability experiment to evaluate the effects of modifying specific attributes, such as skin colour or hair colour. Our generator successfully alters the hair and skin colour of faces within its latent space, and achieves higher control over hair colour than \cite{kowalski2020config} on the generated images. Figures \ref{fig:front} and \ref{fig:generateds} show the qualitative results of controllability for these attributes.

To quantitatively assess the controllability of our framework, we follow \cite{kowalski2020config} and randomly select 1000 images \(I_test\) from the FFHQ validation set, encode them into the latent space \(z = E_F(I_test)\), and then exchange the latent factor \(z_i\) associated with a specific attribute \(v\) (such as hair colour) with a factor obtained from \(E_C\). For each attribute \(v\), we generate two images: \(I^+\) where the attribute is set to a value \(v^+\) (e.g., blonde hair), and \(I^-\) where the attribute takes a semantically opposite value \(v^-\) (e.g., black hair). This results in pairs of images \((I^+, I^-)\) that should be nearly identical except for the selected attribute \(v\), highlighting the differences. To measure these differences, we employ an attribute predictor denoted as \(C_{\text{pred}}\). We train a MobileNet v2 architecture on skin and hair colour, leveraging attribute labels and images from \cite{yucer2022measuring}, and validate it on \(I_test\). In an ideal scenario, \(C_{\text{pred}}(I+)\) should be 1, \(C_{\text{pred}}(I-)\), and the Mean Absolute Difference (MD) for other facial attributes should converge to 0. 

\begin{figure*}[ht]
\begin{center}
\includegraphics[width=\textwidth]{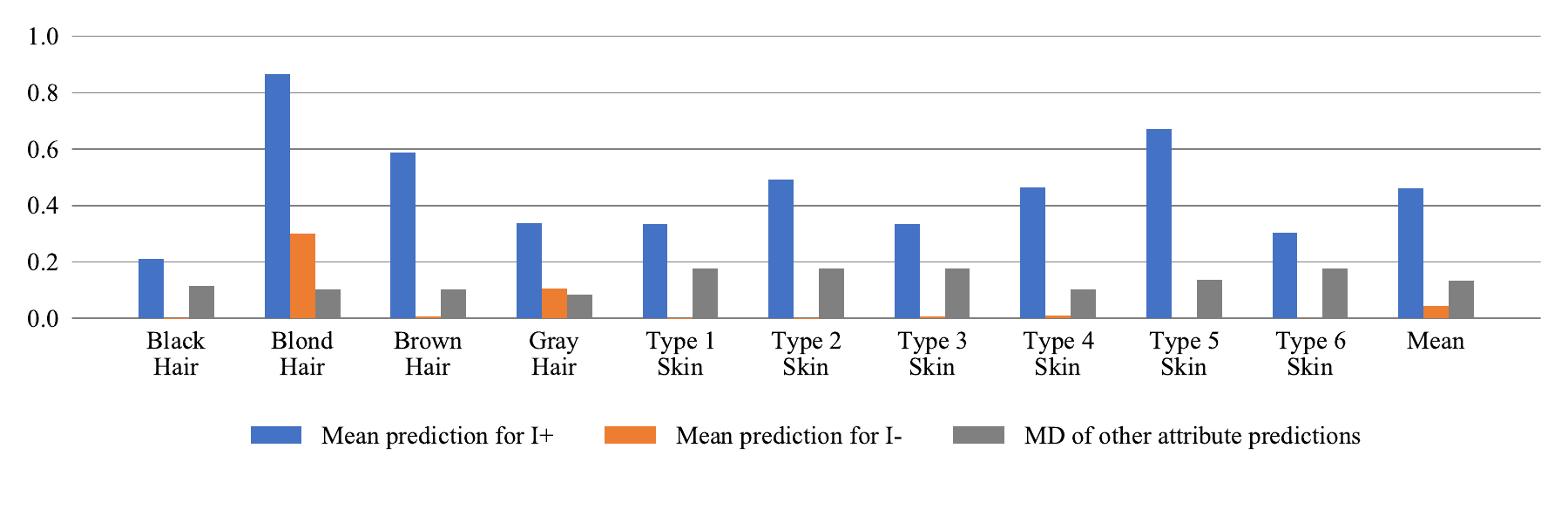}
\end{center}
\vspace{-0.7cm}
\caption{Evaluation of control and disentanglement ability of our proposed framework. Blue and orange bars represent attribute values for images with the respective attribute (\(I+\) for higher values, \(I-\) for lower values). Gray bars indicate differences in other attributes (MD and \(C_{diff}\) for lower values).}

\label{fig:control}
\end{figure*}
Figure \ref{fig:control} illustrates that \(C_{\text{pred}}(I+)\) is generally greater than \(C_{\text{pred}}(I-)\), while the MD for other attributes remains near 0. The highest controllability is observed for skin type 5 and blond and brown hair attributes, where \(C_{\text{pred}}(I+)\) approximates the ideal value of 1. In contrast, the lowest level of control is observed for skin type 1 and black hair attributes. These substantial discrepancies arise from the attribute prediction model capacity on such attributes, as it is trained on VGGFace2 dataset \cite{cao2018vggface2}, which contains a notably low count of Type 1 instances (as indicated by the distribution in \cite{yucer2022measuring}). Consequently, we achieve superior control over hair colour attributes in comparison to \cite{kowalski2020config}, the only possible identical attributes available for comparison.

\begin{figure}[!htbp]
\centering
    \includegraphics[width=\linewidth]{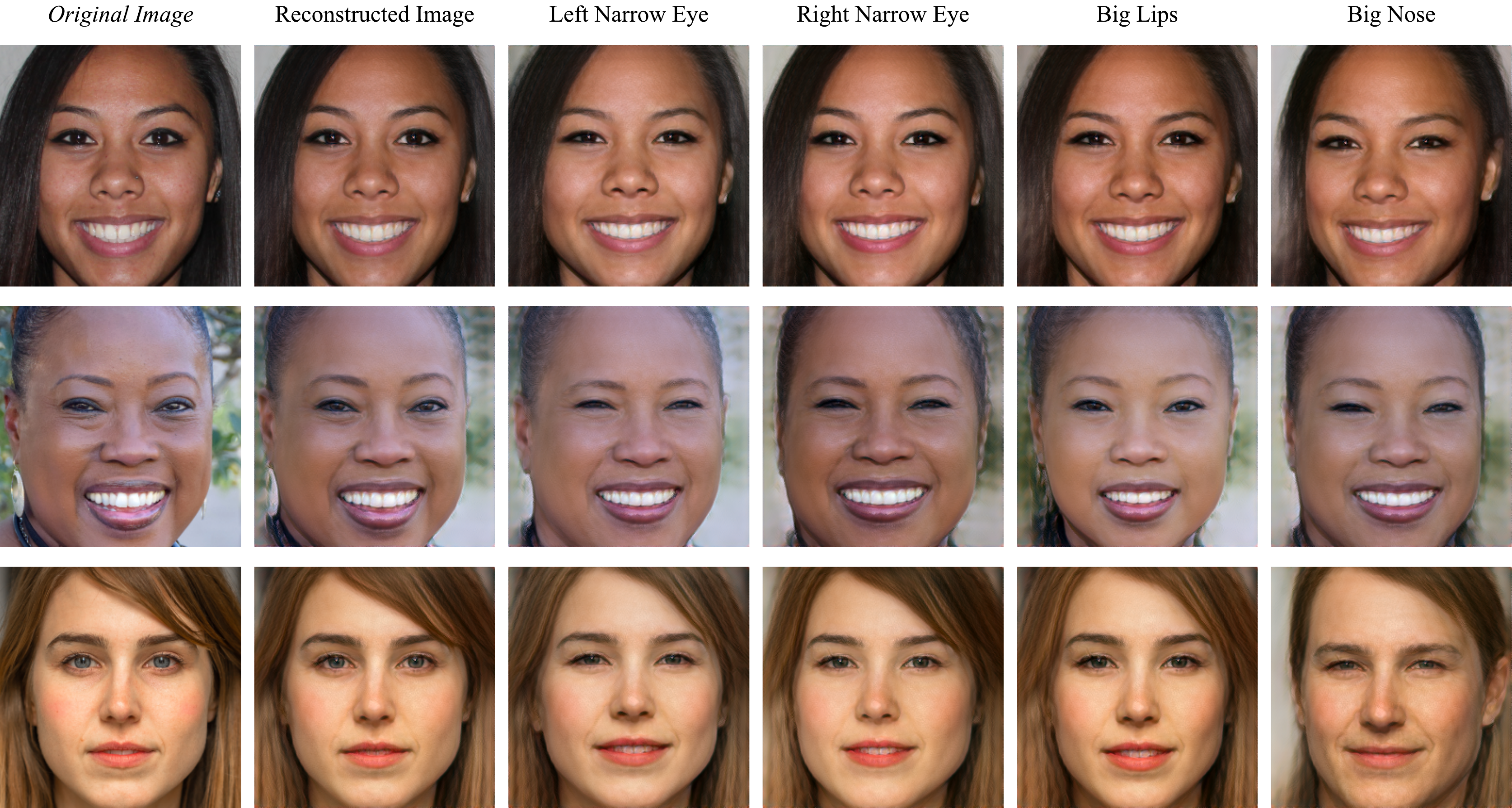}
    \captionof{figure}{Failure modes. Eye Shape Control: leads to a slight appearance shift, affecting both eyes simultaneously. Nose and Lips Control: results in change of unrelated attributes such as pose and mouth openness.}
    \label{fig:failuremode1}
    
\end{figure}

Conversely, our framework encounters challenges in disentangling nose, eye, and mouth shapes. For instance, interchanging left-right eyes leads to alterations in the shape of both eyes. Moreover, altering the nose or lips causes changes in the facial pose and shape. The failure modes of these shape-related attribute changes are presented in Figure \ref{fig:failuremode1}. In the left or right narrow eye control, our framework exhibits two common issues: firstly, it tends to simultaneously alter both eyes or neither, and secondly, it misinterprets narrow eyes as closed eyes in some cases, as seen in the middle row of Figure \ref{fig:failuremode1}. Similarly, for controlling the nose and lips attributes, we observe entanglement with unrelated factors such as pose and mouth openness, as presented in Figure \ref{fig:failuremode1}. We hypothesise that adopting an enhanced feature representation models, such as visual transformers \cite{dosovitskiy2020image} applied to manually generated patch imagery, could lead to substantial improvements in our ability to disentangle these facial features effectively.

\section{Discussion}

\noindent
\textbf{Importance of Training Distribution of Generative Models:} Race-related phenotype disentanglement through generative processes can address racial bias and provide deeper insights into the underlying reasons for disparate performances within racial groupings. However, GANs \cite{maluleke2022studying} reflect the
discrepancies of the training data in the synthesised outputs. Despite our efforts with the CelebA-HQ-Clean-Augmented dataset to reduce the influence of imbalanced distribution of training data on GANs, some unintended correlations still appear. Specifically, when our model was fine-tuned to modify skin colour, it displayed an unintended correlation: associating darker skin tones with eyeglasses (likely due to numerous eyeglass samples within FFHQ) and blonde hair with femininity (17\% of the CelebHQ samples were women with blonde hair). 
\\
\noindent
Additionally, we noted challenges in controlling darker skin tones compared to lighter skin tone ones, possibly due to the symmetric algorithmic bias arises when the imbalances in the training data are magnified in the generated data \cite{maluleke2022studying}.
\\
\noindent
\textbf{Comparison of Entanglement for Shape and Colour Parameters:}
Achieving explicit control over shape-related parameters is more challenging than colour-related ones. This difficulty could arise from inadequate representation of shape features or the greater entanglement of shape with identity, or limitations of StyleGAN2 in handling shape information. Failure modes of such attribute parameter change are illustrated in the Supplementary Material.

\section{Ethical Considerations}

\noindent
\textbf{Use of Face Datasets:} We conduct our experiments using face datasets including CelebA-HQ \cite{karras2017progressive}, FFHQ \cite{karras2019style} which are publicly available for research use only. The reader is directed to the original source publication and the associated research organisation for access to these datasets. 
\\
\noindent
\textbf{Face Editing and Generation:} Our main purpose in synthesising face imagery is to reduce the perpetuation of racial bias caused by imbalanced distributions in face recognition datasets and enable deeper level of analysis of racial bias. To avoid the potential misuse of the synthesised images, we have decided not to publicly share the generated data. However, it may be available upon request for research purposes.

\section{Conclusion}
\noindent
In this study, we introduce a framework, building upon ConfigNet, that disentangles race-related facial phenotypes in a latent space. Our approach leverages 2D publicly available datasets and employs straightforward 2D handcrafted metrics for latent space factorisation. We achieve fine-grained control over racial phenotypes with improves photorealism and controllability compared to ConfigNet without requiring any synthetic data. Although the disentanglement of certain identity-relevant attributes was not entirely controllable, we believe improved and more representative feature metrics will address this in the future. 
\\
\noindent
Future work will follow our primary purpose which is to aid future research on racial bias, as our network facilitates the generation of race-related facial appearance variations and a disentangled feature space. To the best of our knowledge, our study is the first to attempt disentangling and exerting explicit control over such crucial race-related facial phenotype, paving new avenues for evaluating racial bias in automated facial analysis tasks.

    \clearpage
{\small
\bibliographystyle{ieee_fullname}
\bibliography{egbib}
}
\end{document}